\documentclass[letterpaper]{article} 
\usepackage{aaai2026}  
\usepackage{times}  
\usepackage{helvet}  
\usepackage{courier}  
\usepackage[hyphens]{url}  
\usepackage{graphicx} 
\usepackage{natbib}  
\usepackage{caption} 
\usepackage{colortbl}

\usepackage{algorithm}
\usepackage{algorithmic}
\usepackage{subfig}
\usepackage{booktabs}
\usepackage{amssymb}

\usepackage{newfloat}
\usepackage{listings}
\usepackage{placeins}
\usepackage{amsmath}
\DeclareCaptionStyle{ruled}{labelfont=normalfont,labelsep=colon,strut=off} 
\floatstyle{ruled}
\newfloat{listing}{tb}{lst}{}
\floatname{listing}{Listing}
\title{FoodRL: A  Reinforcement Learning Ensembling Framework \\For In-Kind Food Donation Forecasting}
\author{
    Esha Sharma\textsuperscript{\rm 1},
    Lauren Davis\textsuperscript{\rm 2},
    Julie Ivy\textsuperscript{\rm 3},
    Min Chi\textsuperscript{\rm 1}
}
\affiliations{
    \textsuperscript{\rm 1}Department of Computer Science, North Carolina State University, Raleigh, NC, USA\\
    \textsuperscript{\rm 2}Department of Industrial \& Systems Engineering, North Carolina A\&T State University, Greensboro, NC, USA\\
    \textsuperscript{\rm 3}Department of Industrial \& Operations Engineering, University of Michigan, Ann Arbor, MI, USA\\
    \{esharma2, mchi\}@ncsu.edu,\; lbdavis@ncat.edu,\; jsivy@umich.edu
}

\begin{document}

\maketitle

\begin{abstract}

Food banks are crucial for alleviating food insecurity, but their effectiveness hinges on accurately forecasting highly volatile \textbf{in-kind donations} to ensure equitable and efficient resource distribution. Traditional forecasting models often fail to maintain consistent accuracy due to unpredictable fluctuations and concept drift driven by seasonal variations and natural disasters such as hurricanes in the Southeastern U.S. and wildfires in the West Coast. To address these challenges, we propose \textbf{FoodRL}, a novel \textit{reinforcement learning (RL)} based metalearning framework that clusters and dynamically weights diverse forecasting models based on recent performance and contextual information. Evaluated on multi-year data from two structurally distinct U.S. food banks-one large regional West Coast food bank affected by wildfires and another state-level East Coast food bank consistently impacted by hurricanes, FoodRL consistently outperforms baseline methods, particularly during periods of disruption or decline. By delivering more reliable and adaptive forecasts, FoodRL can facilitate the redistribution of food equivalent to 1.7 million additional meals annually,  demonstrating its significant potential for social impact as well as adaptive ensemble learning for humanitarian supply chains.

\end{abstract}


\section{Introduction}
Food insecurity, as defined by the USDA, refers to limited or uncertain access to adequate food for an active and healthy life \cite{coleman2018}. Food banks play a central role in mitigating this issue by distributing surplus food through charitable networks \cite{berner2004shifting}. In 2018, over 37 million Americans, including more than 11 million children, were affected by food insecurity \cite{coleman2018, feedingamerica}, and the COVID-19 pandemic further exacerbated this crisis \cite{brocksurveycovid}. Food banks become especially critical during natural disasters, such as hurricanes in the Southeastern U.S. and wildfires on the West Coast. In these periods, demand for food assistance spikes while supply, particularly from in-kind donations, often drops. These disruptions magnify operational challenges, making accurate forecasting essential for timely distribution, effective resource allocation, and reduced waste. Donations made to food banks generally arrive in two distinct forms: \textbf{monetary} and \textbf{in-kind}. Monetary support, often sourced from state or federal government funds, provides food banks with operational flexibility and ease of allocation.  In-kind donations, however,  present a more complex challenge as they rely on voluntary contributions from individuals, retailers, and organizations, which are often irregular and thus hard to forecast \cite{orgut2016achieving, davis2016analysis}. Forecasting is further complicated by external factors such as market dynamics, donor sentiment, and large-scale disruptions \cite{bennett2003factors, sargeant1999charitable, schlegelmilch1997characteristics, verhaert2011empathy, oloruntoba2006humanitarian, van2006humanitarian}. These challenges make it difficult to anticipate donations which in turn makes it difficult to plan distribution routes and allocate storage, especially during disasters, such as hurricanes in the Southeastern U.S. and wildfires in the West Coast, when in-kind donations often decline. This creates serious challenges in ensuring equitable and efficient resource delivery at critical times \cite{bazerghi2016role, davis2016analysis}. Prior forecasting efforts in this domain have primarily used classical time-series models such as Autoregressive Integrated Moving Average (ARIMA), Exponential Smoothing with Error, Trend, and Seasonality (ETS), and Support Vector Regression (SVR) \citep{davis2016analysis, pugh2017forecast, nuamah2015predicting}. While more recent work has evaluated advanced models such as Long Short Term Memory (LSTM) and Bayesian Structural Time Series (BSTS) \citep{sharma2021data}, metalearning approaches, including reinforcement learning (RL), have not yet been explored in this context. In this work, we analyze multi-year food donation data from two food banks to identify evidence of concept drift, where shifts in the data distribution reduce model performance over time \cite{widmer1996learning, lu2018learning}. Using k-means clustering, we capture recurring drift patterns linked to real-world disruptions such as COVID-19 and natural disasters. Prior research shows that ensemble learning can mitigate concept drift by ensembling models based on recent accuracy \citep{schlimmer1986incremental, tsymbal2004problem, widmer1996learning}, and that model diversity improves adaptability in volatile environments \citep{minku2009impact, oliveira2015ensembles}.

We propose \textbf{FoodRL}, a reinforcement learning based metalearning framework that adaptively ensembles forecasts from a large and growing pool of base models. FoodRL clusters similar models to reduce the action space and mitigate overfitting, allowing the RL agent to assign dynamic weights that respond to evolving data conditions. We apply this method to in-kind donation data from two large food banks, one serving 34 urban and rural counties on the East Coast (EFB), and the other serving a dense urban region on the West Coast (WFB). Despite their differences in geography, cost of living, and donor behavior, both face persistent food insecurity and frequent disruptions \cite{worldpopulationreview, nerdwallet}. Crucially, the types of disasters each food bank faces differ: EFB experiences hurricanes, while WFB experiences wildfires, mudslides, and earthquakes. These events often trigger abrupt changes in in-kind donations, making accurate, adaptive forecasting especially important. Our work directly addresses this challenge by building a forecasting framework that accounts for these disruptions and supports real-time decision-making under uncertainty. Reinforcement learning offers a natural fit by framing ensemble weighting as a sequential decision-making problem. The RL agent learns to adjust model weights over time based on forecast accuracy, making it well-suited for non-stationary environments \cite{sutton2018reinforcement, fu2022reinforcement}. Our results show that FoodRL consistently outperforms these baseline models, particularly under concept drift conditions involving extreme declines and subtle trends. To our knowledge, this is the first application of reinforcement learning based metalearning to the problem of food donation forecasting.

\textbf{Contributions of this work:}
\begin{itemize}
    \item \textit{Meta-learning for Ensembling Food Bank Predictions:} We compare three metalearning techniques, Simple Averaging, Genetic Algorithms, and Reinforcement Learning, at both weekly and monthly levels across two structurally distinct food banks.
    \begin{itemize}
        \item \textit{Genetic Algorithm:} First use of Genetic Algorithm (GA) for metalearning for food donation forecasting.
        \item \textit{Reinforcement Learning:} RL is applied to adaptively combine model outputs based on recent performance.
        \item \textit{Food RL:} A new RL-based ensemble tailored to volatile, sparse food donation data with clustered action spaces.
    \end{itemize}
    \item \textit{Categorizing Concept Drift in Food Bank Data:} We identify and cluster recurring drift patterns to better understand performance degradation.
    \item \textit{Identifying Optimal Meta-learning Models for Different Drift Types:} We assess which ensemble strategies perform best under specific drift conditions.
\end{itemize}

\section{Related Work}

\subsection{Prior Work on Food Donation Forecasting} 

Previous research has explored a range of forecasting approaches for food bank donations, primarily relying on classical time-series models such as ARIMA, MA, ETS, and SVR \cite{davis2016analysis, pugh2017forecast, nuamah2015predicting, brock2012approach}. For example, Davis et al.\ evaluated ARIMA, MA, and ETS using six years of monthly donation data from a single food bank and found ETS yielded the lowest Mean Absolute Percentage Error (MAPE) \cite{davis2016analysis}. Pugh and Davis later extended this work by incorporating SVR and demonstrated its superior performance over ETS using a nine-year dataset \cite{pugh2017forecast}. More recent work examined in-kind donation forecasting at both weekly and monthly intervals, comparing traditional models with more advanced techniques such as LSTM and Bayesian Structural Time Series (BSTS) across a range of training horizons and window strategies. Results indicated that no single model, training length, or window approach consistently outperformed others across food banks or training data length \citep{sharma2021data}. Another line of research focused on donor-level modeling. For instance, forecasts for donations were improved by first clustering donors based on six characteristics: reliability, service score, product variety, quantity received, wastage percentage, and donor affiliation-and then generating ensemble forecasts from five models using aggregated monthly donation amounts \cite{paul2022ensemble}. In contrast to these approaches, our work shifts focus toward dynamic model ensembling at both weekly and monthly levels. We directly learn  ensemble weights from time-series patterns and concept drift using metalearning strategies.

\subsection{The Necessity of Ensembling Learners to Address Concept Drift} 

Concept drift refers to shifts in the predicted variable over time due to previously unseen factors, even if the predictive variables remain unchanged \cite{widmer1996learning, lu2018learning}. In time-series forecasting, this degrades the relevance of past observations. For example, past stock market data may become irrelevant due to sudden unforeseen changes \cite{oliveira2017time}. Donations analysed in our study come from numerous independent donors with highly volatile, irregular, and voluntary patterns, introducing significant randomness \cite{paul2022ensemble, brock2015estimating, sharma2021data}. In time series, concept drift causes values to shift even suddenly, often triggered by sudden events like hurricanes. Our analysis identified several such shifts in food bank data, classified by severity (e.g., extreme, moderate, or slight), helping explain results from previous research that showed that no model works best across all scenarios \cite{sharma2021data}. Studies have shown that model diversity and ensemble strategies can enhance adaptability in dynamic settings \cite{minku2009impact, oliveira2015ensembles}. Minku et al.\ found that high-diversity ensembles handle abrupt, severe, non-repetitive changes better, while low-diversity ones are more stable for gradual or recurring drift \cite{minku2009impact, minku2011ddd}. Other research has shown that ensemble learning methods can effectively address concept drift by dynamically adjusting model weights based on recent performance \cite{schlimmer1986incremental, tsymbal2004problem, widmer1996learning}. Liu et al.\ \cite{liu2023handling} showed that error-based weighting improves performance. Motivated by these insights, we explored an RL based meta-learning approach that dynamically combine learners trained across different model types, window strategies, and training lengths. These ensembles adjust weights based on error, improving adaptability across varying drift conditions in food donation data.

\subsection{Metalearning for Time Series Models} 

Metalearning refers to the process of learning how to select the best algorithm for a given problem \cite{rice1976algorithm}. In data mining, it involves using metadata, such as dataset characteristics and historical model performance, to choose or combine learners effectively \cite{brazdil2008metalearning, prudencio2004meta}. Several studies have applied metalearning to time series forecasting by extracting time series features to select or weight models, demonstrating strong predictive performance \cite{talagala2023meta, montero2020fforma, gastinger2021study, lemke2010meta}. For instance, Montero et al.\ \cite{montero2020fforma} generated weights for model ensembling by optimizing forecasting error using time series features, yielding improved accuracy. While model weighting can mitigate localized error and enhance robustness in concept-drift-prone settings \cite{weigel2008can, minku2009impact}, assigning optimal weights is challenging in such environments \cite{fu2022reinforcement}.
Based on this, we extracted time series features and use reinforcement learning (RL) to dynamically assign weights to models built with varying configurations at each time step. This allows the ensemble to adapt to changing data distributions in food donation patterns by updating weights in response to performance. 

\subsubsection{Simple Averaging}

A common metalearning baseline is Simple Averaging (SA) or computing the mean of predictions from multiple models \cite{clemen1989combining}. This approach is widely used due to its simplicity and robust performance, serving as a benchmark in prior forecasting studies \cite{jose2008simple, makridakis2018m4, montero2020fforma, genre2013combining}. Accordingly, we include average prediction as one of our baselines.

\subsubsection{Genetic algorithm}
Genetic Algorithms (GAs) 
\cite{holland1992adaptation} are optimization techniques inspired by natural selection and have been widely applied in tasks such as model optimization \cite{reif2012meta, friedrichs2005evolutionary, liu2015new, zhong2017genetic, cortez2001meta}, dataset partitioning \cite{cai2013novel, armano2005hybrid}, feature selection \cite{eads2002genetic, frohlich2003feature, gonzalez2015ensemble}, and parameter tuning \cite{gonzalez2015ensemble}. This adaptability makes GAs well-suited for volatile time-series data, such as financial or food donation streams, which exhibit high noise, randomness, and concept drift. Traditional learning methods often struggle with such instability, whereas GAs can search broadly and adapt over time. For example, Pulido et al.\ applied GAs to ensemble neural networks for financial forecasting on the Mexican Stock Exchange, achieving superior performance by evolving combinations based on prediction error \cite{pulido2013genetic}. Similarly, we apply GAs to find optimal model weights for our donation time series, where each learner is defined by a unique combination of model type, data length, and window type. In related work, Wang et al.\ combined ARIMA, exponential smoothing, and neural networks for stock prediction, using GAs to assign weights based on error and improve forecasting in volatile settings \cite{wang2012stock}. Our donation data, which is similarly affected by frequent drift and uncertainty, benefits from this flexible ensembling. 

\subsubsection{Reinforcement Learning Ensembling Approach}
Reinforcement Learning (RL) offers a flexible solution to adapt to frequent concept drift by modeling ensemble weighting as a sequential decision-making task. Here, an agent learns to update model weights over time based on reward signals tied to forecasting accuracy, allowing adaptive learning as new data arrives \cite{sutton2018reinforcement, fu2022reinforcement}. Feng and Zhang \cite{feng2019reinforcement, feng2019reinforced} previously applied Q-learning to select the best forecasting model at each time step, achieving performance gains of up to 50\% over existing methods.
We initially experimented with model selection approaches but found this to be prone to overfitting, particularly as the number of base models grows over time (from 252 to 294 and increasing annually). This expanding model space makes selection approaches increasingly unsuitable. As a result, we focus on model weighting to construct a robust, heterogeneous ensemble that is better suited to handling concept drift in volatile environments. Previously, Fu et al.\ \cite{fu2022reinforcement} demonstrated that reinforcement learning can effectively assign dynamic weights to forecasting models, improving ensemble performance in non-stationary environments. Their method, RL-based model combination (RLMC), uses a classifier to estimate the probability of each model being optimal and then applies a softmax function to derive weights. However, in our setting, where the number of base models ranges from 252 to 294 and increases annually, this approach becomes impractical. 

In contrast, our proposed framework, Food-RL, learns weights from scratch and is specifically designed to handle large, growing sets of models. Rather than assigning weights to individual models, Food-RL clusters similar models to reduce action space complexity. This is particularly important given our dataset’s limited size and the high correlation among models, which would otherwise lead to redundant, high-dimensional action spaces that hinder learning. By leveraging clustering, Food-RL enables more efficient and robust learning in data-scarce, drift-prone environments.

\section{Methods}
\label{sec3}

\noindent\textbf{Problem Formulation}

We solve a time-series prediction problem, where the objective is to ensemble forecasted future donation volumes. These forecasts are generated using historical measurements. 
The dataset for forecasting is defined as:

\[
\textbf{Y} = \{ {y}_1, {y}_2, ..., {y}_N \}
\]

where $N$ denotes the number of time steps. For instance, in the EFB dataset, $N = 131$ for monthly events. Each event at time step $t$ is associated with a multivariate input vector $\boldsymbol{x}_t \in \mathbb{R}^D$, where $D$ is the number of features of the time series and donation-related data. The forecasting task is to estimate the next-step donation volume $y_t$, given the historical input trajectory up to time $t$, i.e., $\{{x}_1, {x}_2, ..., {x}_t\}$. The corresponding prediction is denoted as $\hat{y}_t$, and the ground truth donation at time $t+1$ is $y_t$. This work leverages multiple forecasts each generated using a predictive model each trained to learn the mapping from historical input sequences to next-step donation values but with a variation in the model, length, and window as noted in previous work by Sharma et. al \cite{sharma2021data}. The final prediction $\hat{y}_t$ is obtained by ensembling the predictions from these individual models. For each time-step ${t}$, we have a set of $k$ predictions  $\boldsymbol{\hat{y}_t^{(k)}}$ generated where $k$ varies with the time-step. The goal of this work is to assign a weight $wt_t^{(i)}$ to each $\hat{y}_t^{(i)}$ to find the final weighted prediction ${y}^\bigtriangleup_t$ at time-step $t$.
For each time step $t$, we generate a set of $k_t$ predictions from different models or sources:

\[
\hat{\boldsymbol{y}}_t = \{ \hat{y}_t^{(1)}, \hat{y}_t^{(2)}, \dots, \hat{y}_t^{(k_t)} \}
\]

where $\hat{y}_t^{(i)}$ denotes the $i^\text{th}$ prediction available at time $t$, and $k_t$ represents the number of predictions available at that specific time step (which may vary across time).

The final prediction is computed using an ensemble function $\mathcal{E}_t(\cdot)$ applied over the set of predictions at time $t$:

\[
\hat{y}^\bigtriangleup_t = \mathcal{E}_t \left( \hat{y}_t^{(1)}, \hat{y}_t^{(2)}, \dots, \hat{y}_t^{(k_t)} \right)
\]

\emph{\textbf{Baseline: Average of Predictions}} 

The baseline $\mathcal{E}_t(\cdot)$ is implemented as an average of the $k_t$ predictions, where each prediction $\hat{y}_t^{(i)}$ is assigned an equal weight $w_t^{(i)}$ such that $\sum_{i=1}^{k_t} w_t^{(i)} = 1$:

\[
\hat{y}^\bigtriangleup_t = \sum_{i=1}^{k_t} w_t^{(i)} \cdot \hat{y}_t^{(i)}
\] 

\emph{\textbf{Baseline: Genetic Algorithm}} 

In this work, we utilize a GA \cite{holland1992adaptation} to optimize the set of weights assigned to the individual learners in the ensemble. The GA evolves candidate solutions through iterative applications of three primary operations: \emph{Selection}, \emph{Crossover}, and \emph{Mutation}. Initially, a population of $N$ individuals is randomly generated, where each individual represents a potential set of weights $\{ w_t^{(1)}, w_t^{(2)}, \dots, w_t^{(k_t)} \}$ to be assigned to the $k_t$ predictions at time step $t$. The fitness of each individual is evaluated using a predefined error metric (e.g., Mean Squared Error), which quantifies the prediction error when using the candidate weights in the ensemble. The GA then iteratively evolves the population through the following steps:

\begin{itemize}
    \item \textbf{Selection:} A subset of individuals with the best fitness scores is selected to form the mating pool.
    \item \textbf{Crossover:} Pairs of selected individuals are recombined using a crossover probability to generate new candidate solutions by mixing their weights.
    \item \textbf{Mutation:} With a given mutation probability, small random changes are introduced into the weights of selected individuals to maintain genetic diversity and explore new regions of the solution space.
\end{itemize}

This evolutionary process is repeated for a fixed number of generations or until convergence. The best-performing individual from the final generation is then selected as the optimal weight vector and used in the ensemble model for computing the final weighted prediction:

\[
\hat{y}^\bigtriangleup_t = \sum_{i=1}^{k_t} w_t^{(i)} \cdot \hat{y}_t^{(i)}
\]

This optimization approach allows the model to dynamically learn time-step-specific weightings of predictions, improving ensemble performance over static or unweighted combinations.\\

\emph{\textbf{Reinforcement Learning}} \\
We define the task of ensembling as the output of a RL agent that assigns dynamic weights to the predictions at time step $t$. Specifically, the RL agent observes the state $\mathcal{S}_t$ and selects an action $\mathcal{A}_t = \{w_t^{(1)}, w_t^{(2)}, \dots, w_t^{(k_t)}\}$ according to a learned policy $\pi$, such that the ensemble function $\mathcal{E}_t(\cdot)$ is defined as the weighted sum of learner predictions using weights from the agent's action in an adaptive, data-driven manner that evolves over time.

\[
\hat{y}^\bigtriangleup_t  = \sum_{i=1}^{k_t} w_t^{(i)} \cdot \hat{y}_t^{(i)}
\]

\paragraph{State}
The state $\mathcal{S}_t$ at each time step $t$ is defined as a feature vector capturing the current context, including:
\begin{itemize}
    \item Temporal, and statistical characteristics extracted from historical observations of the time series,
    \item Predictions from all available learners at time step $t$,
    \item The previous set of weights assigned by the agent at time step $t-1$.
\end{itemize}

\paragraph{Action}
The action $\mathcal{A}_t$ is a vector of weights assigned to the $k_t$ base learners at time step $t$:
\[
\mathcal{A}_t = \{ w_t^{(1)}, w_t^{(2)}, \dots, w_t^{(k_t)} \}
\]
Each weight $w_t^{(i)} \in [0, 1]$, and the weights are normalized such that $\sum_{i=1}^{k_t} w_t^{(i)} = 1$.

\paragraph{Reward}  
The reward \( \mathcal{R}_t \) at time step \( t \) is defined as the negative Mean Absolute Percentage Error (MAPE) between the ensemble prediction \( \hat{y}^\bigtriangleup_t \) and the true value \( y_t \):

\[
\mathcal{R}_t = -\text{MAPE}(\hat{y}^\bigtriangleup_t, y_t)
\]
where $\hat{y}^\bigtriangleup_t$ is the ensemble prediction obtained using the current weights.

\paragraph{Transition Dynamics.}
At each time step, the RL agent selects an action (ensemble weights), computes the weighted prediction, observes the reward, and transitions to the next time step. The state is updated with the new weights, features, and predictions. This formulation enables the agent to adaptively refine its ensembling strategy as temporal patterns in the time series evolve.

\paragraph{Objective.}
The goal of the RL agent is to learn a policy $\pi(\mathcal{A}_t | \mathcal{S}_t)$ that maximizes the expected cumulative reward over time:
\[
\max_\pi \mathbb{E} \left[ \sum_{t=1}^{T} \gamma^t \mathcal{R}_t \right]
\]
where $\gamma \in [0, 1]$ is the discount factor. \\

\emph{\textbf{Food RL}} 

We define Food RL as a framework to enable RL to learn better by reducing action space and redundancy among the base learners. To this end, we apply unsupervised clustering using the K-Means algorithm to cluster predictions generated by each model. Let $\mathcal{M} = \{f^{(1)}, f^{(2)}, \dots, f^{(K)}\}$ denote the set of $K$ base learners. We apply K-Means clustering to partition the learners into $C$ distinct clusters:
\[
\mathcal{C}_1, \mathcal{C}_2, \dots, \mathcal{C}_C
\]
We then apply a RL agent to learn a dynamic weighting policy over these clusters. The RL agent observes the state $\mathcal{S}_t$, including time-series statistics, cluster predictions, and previous weight assignments, and outputs a normalized weight vector:
\[
\mathcal{A}_t = \{ w_t^{(1)}, w_t^{(2)}, \dots, w_t^{(C)} \}
\]
Each weight $w_t^{(i)} \in [0, 1]$, and the weights are normalized such that $\sum_{i=1}^{k_t} w_t^{(i)} = 1$. The final ensemble prediction is computed as a weighted sum over the cluster predictions:
\[
\hat{y}^\bigtriangleup_t = \sum_{j=1}^{C} w_t^{(j)} \cdot \hat{y}_t^{(j)}
\]

\section{Experimental Setup}
\label{sec:experiment}

\subsection{Dataset: Constituent Predictions for Metalearning}

For the constituent models, we use the dataset of model predictions introduced by Sharma et al.~\cite{sharma2021data}, which demonstrated that no single model consistently outperforms others across the full time span. This dataset provides monthly-resolution forecasts, where each model represents a unique combination of \textit{model type}, \textit{training window length}, and \textit{window strategy}. It spans 14 years (2007–2020) for EFB and 5 years (2014–2019) for WFB, with donation volumes aggregated in pounds. As more training data becomes available over time, the number of model forecasts per time step increases: up to 252 for EFB and 105 for WFB per month. The model pool includes ARIMA, MA, and LSTM, as well as two variants each of BSTS (with and without covariates) and ETS (with and without trend and seasonality) \cite{sharma2021data}.

\subsection{Goal}
The goal of our experiments was to find a method to effectively ensemble precdictions generated using different combinations of model, window, and length to minimise the error at each time step $t$ 
\[
e_t = \hat{y}^\bigtriangleup_t  - {y}_t 
\]

\subsection{Feature Engineering}
\label{sec:fe}

We utilized Python's TSFEL library \cite{barandas2020tsfel} to extract time series features for both monthly and weekly forecasting models. For the monthly level, features were generated using rolling windows of 2, 4, 6, and 12 months; for the weekly level, we used 2, 6, 12, 26, 38, and 52 weeks. These varying window lengths allow the model to capture both short-term fluctuations and long-range seasonal patterns. At each time step $t$, we constructed a feature vector using only historical data up to $t-1$, i.e., the rolling window spanned the interval $[t-w, t-1]$, where $w$ is the window size. These features include both temporal descriptors (e.g., slope, mean difference, median difference) and statistical summaries (e.g., maximum, mean, standard deviation). Our framework uses these features to ensemble predictions for donations at time $t$ while ensuring strict one-step-ahead forecasting: all features are derived solely from past data, with no leakage from the future, that is, for each time step $t$ we used features generated using data until time step $t-1$ for ensembling predictions at time $t$.

\subsection{Training and Testing Data}
For metalearning, the testing data for each dataset comprised the last two years, while the training data included all prior years. 

\subsection{Building Models for Prediction}

\subsubsection{\textbf{GA Approach}}
To learn adaptive ensemble weights for model predictions at each time step, we employ a GA-based optimization approach. At each forecasting step $t$, we construct a sliding window of historical data with a fixed length of 24, using it to train the ensemble. Within this window, we identify all models that have provided non-missing predictions and use their outputs to form the training set. The GA is configured to optimize a vector of weights corresponding to the available models, with the goal of minimizing the Mean Absolute Error (MAE) between the weighted ensemble prediction and the actual donation volumes in the training window. The fitness function is defined as the negative MAE to align with the GA’s maximization objective. We use the pygad library with 50 generations and each gene (model weight) is constrained to the range $[0, 1]$, and the final weights are normalized to ensure they sum to 1. Once the optimal weights are learned, they are applied to the model predictions at time $t$ to generate the final ensemble forecast. This process is repeated for each time step, enabling the ensemble to adapt over time based on recent model performance.

\subsubsection{\textbf{RL Approach}}
To train the RL agent, we use the Proximal Policy Optimization (PPO) algorithm with a multilayer perceptron policy. The agent is trained in a custom Gym environment on historical donation data, where each episode corresponds to a sequence of weekly or monthly time steps. During training, the agent interacts with the environment by selecting ensemble weight vectors based on the current state (which includes model predictions, donor-related features, and previous weights), and receives feedback via a reward signal based on the negative Mean Absolute Percentage Error (MAPE). Once training is complete, the learned policy is evaluated on the test dataset to assess its performance. At each test time step, the trained agent generates ensemble weights, which are applied to the available model predictions to produce a weighted forecast. The predicted values are compared against ground truth donation volumes. 

We also experimented with other algorithms, including Soft Actor Critic (SAC), Twin Delayed Deep Deterministic Policy Gradient (TD3), and Deep Deterministic Policy Gradient (DDPG), but they all exhibited high variability and poor performance. In contrast, Proximal Policy Optimization (PPO) provided the most stable and reliable results, especially after applying clustering

\subsubsection{Food RL Approach}

We follow the same approach as RL, but instead of using individual models, we group the donation models into clusters using training data. We determine the optimal number of clusters for each case: 40 for monthly EFB, 40 for weekly EFB, 60 for monthly WFB, and 60 for weekly WFB. For both the training and testing datasets, we aggregate the predictions of models within each cluster by averaging their outputs at each time step. This results in a lower-dimensional representation of the ensemble, where each new feature represents the average prediction of a cluster rather than an individual model.

\subsection{Implementation Details}
We used the standard PPO implementation from Stable-Baselines3 with an MlpPolicy (two 64-unit hidden layers), a learning rate of $10^{-4}$, and the default 10 optimization epochs per update. The number of clusters (k) was hyperparameter-tuned during training, resulting in a fixed number of clusters which were then fixed.

\subsection{Evaluation Metrics}
To evaluate the performance and compare the models, we employ \emph{Mean Average Error (MAE)} \cite{willmott2005advantages} and  \emph{Mean Average Percentage Error (MAPE)} \cite{karakoyun2018comparison, islam2012empirical}.\\
The \emph{MAE} is defined as:

\emph{MAE} $= (\sum_{i=1}^{t}  abs(\hat{y}^\bigtriangleup_i - y\textsubscript{i})) /t$ 

\noindent The \emph{MAPE} is defined as: 

\emph{MAPE} $= (\sum_{i=1}^{t}  (abs(\hat{y}^\bigtriangleup_i- y\textsubscript{i})/y\textsubscript{i}) * 100) /t$.
\\

\section{Results and Discussion}
\label{sec:resdisc}

\subsection{Concept Drift in Food Bank Data}
\begin{figure*}[t]
    \centering

    \subfloat[EFB \label{fig:cd1}]{
        \includegraphics[width=0.48\textwidth, height=4.5cm]{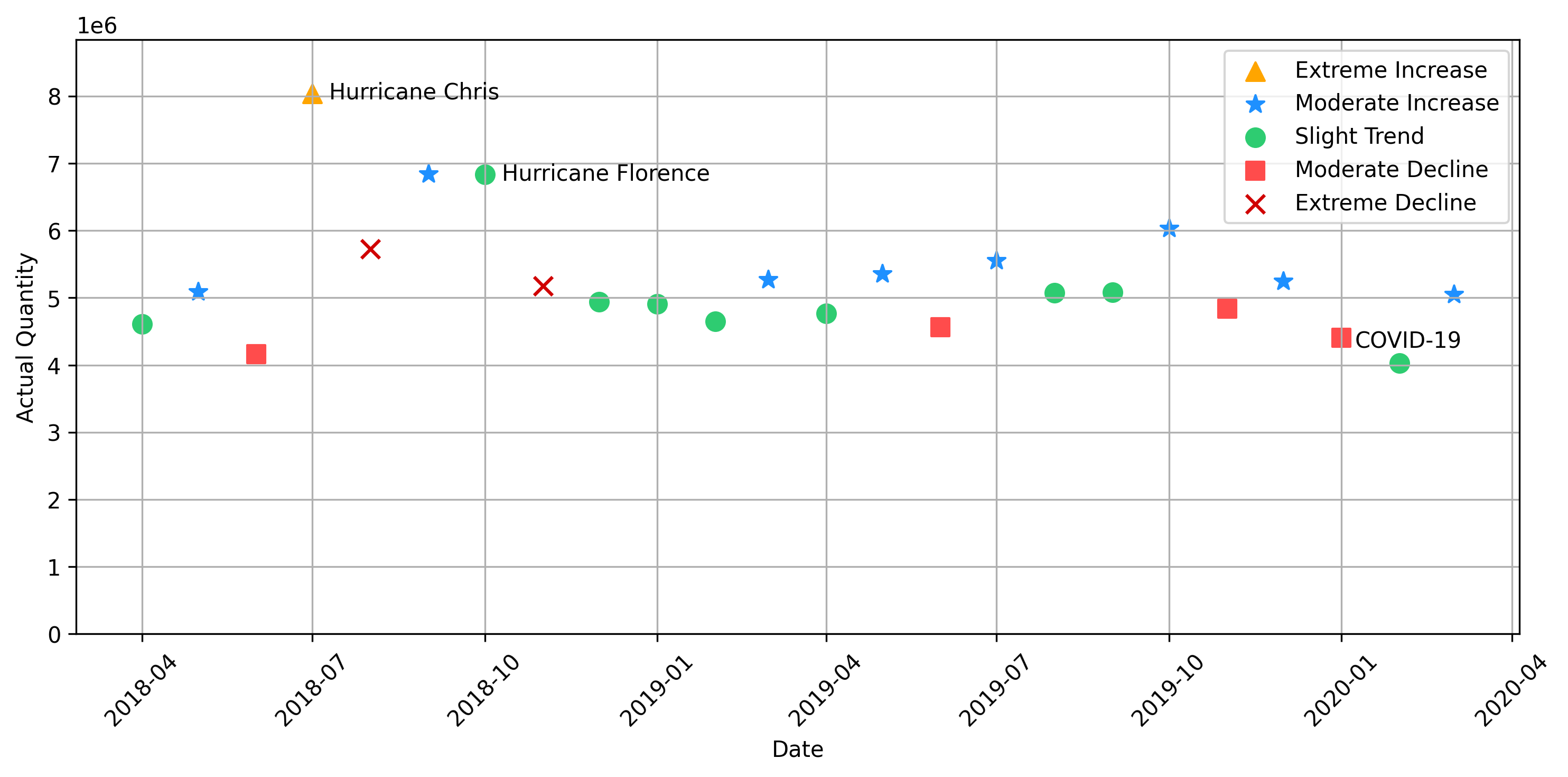}
    }
    \hfill
    \subfloat[WFB \label{fig:cd2}]{
        \includegraphics[width=0.48\textwidth, height=4.5cm]{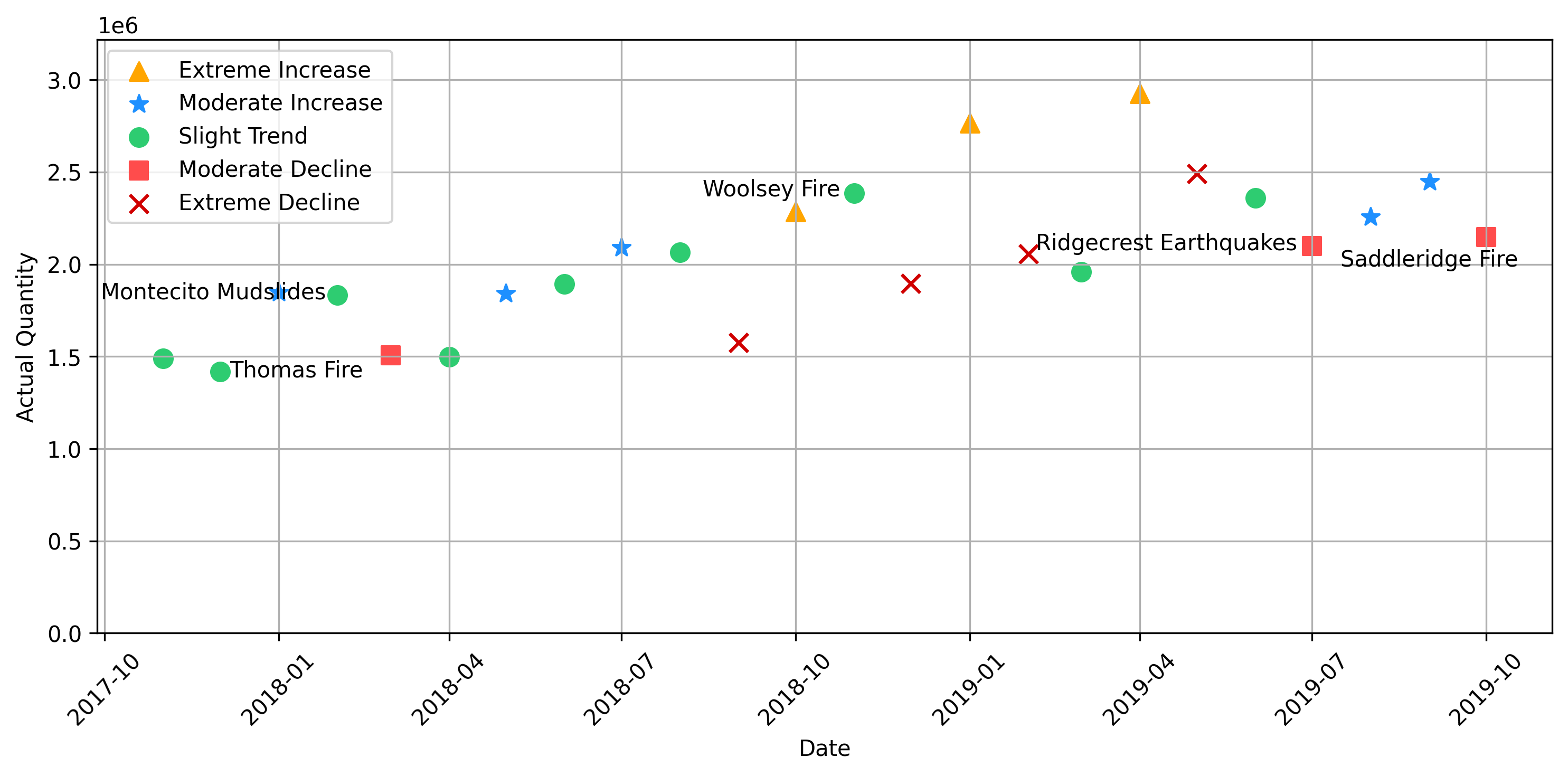}
    }
    \caption{Concept Drift in Food Bank Data }
    \label{fig:cd_all}
\end{figure*}


Upon analysis of the donation data, presented in Figure 1, we observed distinct patterns between EFB and WFB. EFB data spans a wider range, from 4.0 to over 8.0 million pounds, and shows a sharp spike in July 2018, likely linked to Hurricane Chris. In contrast, WFB donations range from 1.4 to 3.0 million pounds and show several, but less extreme, increases around late 2018 to early 2019. To characterize drift, we used K-means clustering, categorizing monthly data into five groups after visual inspection (extreme/moderate increases or declines, slight trend).
EFB data largely consisted of moderate and slight fluctuations, while WFB data displayed more frequent and intense swings, indicating stronger drift. This volatility in WFB aligns with major disasters from 2017–2019, including the Thomas Fire (Dec 2017), Montecito Mudslides (Jan 2018), Woolsey Fire (Nov 2018), Ridgecrest Earthquakes (July 2019), and Saddleridge Fire (Oct 2019) \cite{ladisasters}. In EFB, sharp increases corresponded with hurricanes, notably Hurricane Chris (July 2018) and Hurricane Florence (Sept 2018), one of the state’s most damaging storms in recent years \cite{hurricaneflorence, ncdisasters}.

\begin{table*}[!]
\centering
\caption{Average MAPE by Concept Drift Type}
\label{tab:cluster_mape_double}
\begin{minipage}{0.48\textwidth}
\centering
\textbf{(a) EFB}
\resizebox{\textwidth}{!}{
\begin{tabular}{l|cccc}
\toprule
\textbf{Cluster} & \textbf{SA} & \textbf{GA} & \textbf{RL} & \textbf{Food RL} \\
\midrule
Extreme Increase   & \textbf{41.22} {\scriptsize$\pm$ 0.00}   & 41.58 {\scriptsize$\pm$ 0.00}   & 46.67 {\scriptsize$\pm$ 0.00}   & 41.72 {\scriptsize$\pm$ 0.00} \\
Moderate Increase  & \textbf{7.59} {\scriptsize$\pm$ 6.52}  & 9.28 {\scriptsize$\pm$ 6.58}  & 11.10 {\scriptsize$\pm$ 7.92} & 7.73 {\scriptsize$\pm$ 5.17} \\
Slight Trend       & 10.30 {\scriptsize$\pm$ 6.61}          & 9.36 {\scriptsize$\pm$ 6.40}  & 9.43 {\scriptsize$\pm$ 6.54}  & \textbf{8.82} {\scriptsize$\pm$ 8.30} \\
\rowcolor{red!8}
Moderate Decline   & 13.64 {\scriptsize$\pm$ 2.38}          & 11.17 {\scriptsize$\pm$ 3.67} & 9.86 {\scriptsize$\pm$ 5.04}  & \textbf{9.05} {\scriptsize$\pm$ 4.54} \\
\rowcolor{red!20}
Extreme Decline    & 12.57 {\scriptsize$\pm$ 14.63}         & 13.88 {\scriptsize$\pm$ 11.12}& 12.92 {\scriptsize$\pm$ 6.73} & \textbf{7.65} {\scriptsize$\pm$ 9.21} \\

\bottomrule
\end{tabular}
}
\end{minipage}
\hfill
\begin{minipage}{0.48\textwidth}
\centering
\textbf{(b) WFB}
\resizebox{\textwidth}{!}{
\begin{tabular}{l|cccc}
\toprule
\textbf{Cluster} & \textbf{SA} & \textbf{GA} & \textbf{RL} & \textbf{Food RL} \\
\midrule
Extreme Increase   & \textbf{26.77} {\scriptsize$\pm$ 1.69} & 27.69 {\scriptsize$\pm$ 0.90} & 28.25 {\scriptsize$\pm$ 2.06} & 26.90 {\scriptsize$\pm$ 2.34} \\
Moderate Increase  & \textbf{12.25} {\scriptsize$\pm$ 7.92} & 14.26 {\scriptsize$\pm$ 9.26} & 17.23 {\scriptsize$\pm$ 10.92} & 13.14 {\scriptsize$\pm$ 8.42} \\
Slight Trend       & 8.41 {\scriptsize$\pm$ 3.91}           & 8.80 {\scriptsize$\pm$ 4.25}  & 9.55 {\scriptsize$\pm$ 4.36}  & \textbf{7.75} {\scriptsize$\pm$ 4.44} \\
\rowcolor{red!8}
Moderate Decline   & 15.40 {\scriptsize$\pm$ 3.21}          & 13.86 {\scriptsize$\pm$ 2.11} & \textbf{13.59} {\scriptsize$\pm$ 1.70} & 14.68 {\scriptsize$\pm$ 5.70} \\
\rowcolor{red!20}
Extreme Decline    & 15.82 {\scriptsize$\pm$ 11.11}         & 16.09 {\scriptsize$\pm$ 10.39}& 15.73 {\scriptsize$\pm$ 8.86} & \textbf{14.07} {\scriptsize$\pm$ 8.77} \\
\bottomrule
\end{tabular}
}
\end{minipage}
\end{table*}

\begin{table}[!t]
\centering
\caption{Performance of Metalearning Models}
\label{tab:monthly_only}
\resizebox{\columnwidth}{!}{
\begin{tabular}{l | c c | c c}
\toprule
\textbf{Model} & \multicolumn{2}{c|}{\textbf{EFB}} & \multicolumn{2}{c}{\textbf{WFB}} \\
\cline{2-5}
& MAE ($10^3$lb) & MAPE & MAE ($10^3$lb) & MAPE \\
\midrule
ARIMA        & 741.62\,{\scriptsize$\pm$705.08} & 13.36\,{\scriptsize$\pm$9.80}  & 298.02\,{\scriptsize$\pm$196.83} & 14.33\,{\scriptsize$\pm$7.45} \\
MA           & 750.26\,{\scriptsize$\pm$675.98} & 13.66\,{\scriptsize$\pm$9.44}  & 294.92\,{\scriptsize$\pm$188.46} & 14.20\,{\scriptsize$\pm$8.85} \\
ETS-Plain    & 700.91\,{\scriptsize$\pm$674.27} & 12.69\,{\scriptsize$\pm$9.20}  & 292.56\,{\scriptsize$\pm$223.78} & 13.91\,{\scriptsize$\pm$8.98} \\
ETS-Plus     & 717.57\,{\scriptsize$\pm$663.30} & 13.05\,{\scriptsize$\pm$9.02}  & 300.77\,{\scriptsize$\pm$220.20} & 14.28\,{\scriptsize$\pm$8.75} \\
LSTM         & 750.31\,{\scriptsize$\pm$748.12} & 13.13\,{\scriptsize$\pm$9.06}  & 285.39\,{\scriptsize$\pm$193.79} & 13.72\,{\scriptsize$\pm$7.05} \\
BSTS-Plus    & 706.10\,{\scriptsize$\pm$536.38} & 12.98\,{\scriptsize$\pm$7.18}  & 346.23\,{\scriptsize$\pm$170.63} & 16.73\,{\scriptsize$\pm$7.03} \\
BSTS-Plain   & 677.84\,{\scriptsize$\pm$578.99} & 12.30\,{\scriptsize$\pm$7.55}  & 345.18\,{\scriptsize$\pm$164.90} & 16.71\,{\scriptsize$\pm$6.63} \\
\hline
SA      & 635.41$^{*}$\,{\scriptsize$\pm$672.41} & 11.43\,{\scriptsize$\pm$9.13}  & 283.25$^{*}$\,{\scriptsize$\pm$206.34} & 13.61$^{*}$\,{\scriptsize$\pm$8.27} \\
GA           & 642.15\,{\scriptsize$\pm$676.60} & 11.35$^{*}$\,{\scriptsize$\pm$8.83}  & 296.12\,{\scriptsize$\pm$211.31} & 14.15\,{\scriptsize$\pm$8.46} \\
RL           & 681.72\,{\scriptsize$\pm$762.21} & 11.90\,{\scriptsize$\pm$9.76}  & 314.71\,{\scriptsize$\pm$214.37} & 15.02\,{\scriptsize$\pm$8.61} \\
Food RL & \textbf{564.66}$^{*}$\,{\scriptsize$\pm$686.67} & \textbf{9.77}$^{*}$\,{\scriptsize$\pm$9.03} & \textbf{280.30}$^{*}$\,{\scriptsize$\pm$209.33} & \textbf{13.19}$^{*}$\,{\scriptsize$\pm$8.13} \\
\bottomrule
\end{tabular}
}
\scriptsize{$^*$ Indicates best or second-best model. The best is additionally bolded.}
\end{table}

\subsection{Metalearning Results}
Table~\ref{tab:monthly_only} presents a performance comparison across both food banks. We evaluate FoodRL against two meta-learning baselines: SA and GA, as well as a basic RL ensemble. Additionally, we include constituent forecasting models (ARIMA, MA, ETS, LSTM, and BSTS) previously studied in \cite{sharma2021data}. Food RL achieves the lowest MAPE across EFB and WFB. It outperforms all constituent models as well as baselines, including SA, GA, and standard RL. We conducted the Wilcoxon signed-rank test to assess whether the MAPE of Food RL, differed significantly from that of other models. For EFB, Food RL demonstrated statistically significant improvements over nine models: MA, LSTM, ARIMA, ETS variants, BSTS variants, GA, and SA, with  $p < 0.05$. While Food RL also achieved lower mean MAPE than RL, this difference was not statistically significant. For WFB, Food RL significantly outperformed four models, BSTS variants, RL, and GA with $p < 0.05$. While Food RL also exhibited lower mean MAPE than the remaining models (including SA, ARIMA, ETS variants, and LSTM), these differences were not statistically significant.

Table~\ref{fig:cd_all} presents model performance across different concept drift clusters for both NC and LA. At the monthly level in EFB, RL and Food RL both perform well in capturing declining trends, particularly moderate and extreme declines. Food RL outperforms RL in all these cases. However, all models struggle with extreme increases (MAPE $> 40\% $). These spikes align with hurricane season, July 2018 saw a sharp increase associated with Hurricane Chris, followed by a more significant surge in September due to Hurricane Florence \cite{ncdisasters, hurricaneflorence}. Food RL remains stable during low-volatility periods. A key insight is that both RL-based methods excel at capturing downward trends and adapting to long-term structural changes, such as those caused by COVID-19. Performance differs markedly between EFB and WFB due to variation in drift frequency and severity. WFB shows more frequent extreme declines, while EFB exhibits more moderate declines. For example, WFB’s extreme decline cluster yields the highest errors, yet Food RL consistently performs best, especially on slight trends and extreme declines. In WFB, for moderate trends, Food RL outperforms others; for slight increases, performance is similar across models, with Average slightly ahead. As with EFB, extreme increases remain difficult, with Average performing marginally better. Nevertheless, Food RL proves most robust, excelling in both stable and declining scenarios under high volatility. Overall, extreme increases and declines remain the hardest to predict, particularly in WFB. In summary, while forecasting under concept drift remains difficult, especially for extreme fluctuations, Food RL consistently delivers the lowest MAPE in most settings. Its dynamic, adaptive design enables robust performance across both stable and highly volatile regimes.

\subsection{Ablation Studies}
We compared PPO with TD3, SAC, and DQN. PPO gave us the best results. For instance, PPO outperformed SAC (EFB: 9.77 vs 10.95; WFB: 13.19 vs 13.82). We attribute PPO’s advantage to its clipped surrogate objective, which stabilizes training by constraining large policy updates and reducing variance. This property makes PPO particularly effective in our setting, where data are limited ($<$160 samples) and exhibits high variability. For the reward, we tested MAE, MAPE, SMAPE, as well as an optimized MAPE (comparing to most optimal model) formulations, finding only minor performance differences. Performance improves as more TSFEL features are included, so we do not perform manual feature selection. 

\FloatBarrier
\section{Conclusion}
\label{sec:conclusion}
Accurate forecasting of food donations is vital for minimizing waste and supporting equitable distribution to food-insecure households. However, donation data is highly volatile and non-stationary, making prediction challenging. In this study, we evaluate three meta-learning approaches: Simple Averaging, Genetic Algorithms, and Reinforcement Learning (RL), and introduce Food RL, a RL-based ensemble framework enhanced with clustering to reduce action space and improve adaptability. Across both food banks (EFB and WFB), Food RL achieves the lowest MAPE and MAE in nearly all cases, outperforming all baselines. Clustering improves learning efficiency and robustness under dynamic conditions. To analyze model behavior across time, we segment the data into concept drift patterns. Food RL performs especially well for moderate and extreme declines, as well as slight trends. While all models struggle with extreme increases, often driven by natural disasters, Food RL remains among the top performers. Overall, this work contributes (1) a clustering-based meta-learning method tailored for volatile food donation data, (2) an empirical analysis of concept drift patterns, and (3) evidence of model robustness by drift type. Practically, using Feeding America’s metric of 1.2 pounds per meal \cite{feedingamerica}, Food RL’s improved forecasts translate to roughly 1.66 million additional meals per year for monthly data and 593,000 meals annually at the weekly level.

\section{Impact and Future Work}
\label{sec:fw}

We are currently piloting FoodRL with an industry partner, and both food banks are included in our study. A prototype has been implemented for pilot testing at EFB, and we are actively enhancing and expanding its capabilities. Our forecasts are being integrated into downstream tools for distribution and supply chain management, with initial deployment focused on monthly forecasting to support donation planning and resource allocation during high-risk periods such as the COVID-19 pandemic and hurricane seasons.

For future work, we aim to enhance performance during extreme increase scenarios and improve forecasting accuracy at the weekly level. We also plan to incorporate more granular data to further refine the model. 

While our current focus is on food donation data, the FoodRL framework is broadly applicable to other domains that face similar challenges of volatility, sparsity, and concept drift, such as humanitarian logistics during disasters.

\bibliography{aaai2026}

\FloatBarrier 

\end{document}